\documentclass{article}
\usepackage[table]{xcolor}
\usepackage{enumitem}
\usepackage{graphicx}
\usepackage{tikz}
\usepackage{xspace}
\usepackage{subcaption}
\usepackage{float}
\usepackage{placeins}
\usepackage{indentfirst,csquotes}
\usepackage[backend=biber,style=numeric,sorting=none,natbib=true]{biblatex}
\usepackage{amsmath}
\usepackage{amssymb}
\usepackage{mathtools}
\usepackage{amsthm}
\usepackage{microtype}
\usepackage{booktabs} 
\usepackage{multirow}
\usepackage{array}
\usepackage{url}
\usepackage{algorithm}
\usepackage{algpseudocode}
\usepackage{hyperref,titlesec,fancyhdr,etoolbox}

\definecolor{best}{RGB}{255,220,220}      
\definecolor{second}{RGB}{255,240,200}    

\titleformat{\section}[block]
  {\normalfont\large\bfseries} 
  {\thesection.}               
  {0.6em}                      
  {}

\titlespacing*{\section}{0pt}{2ex plus 1ex minus .2ex}{1.5ex plus .2ex}

\hypersetup{ colorlinks=true, linkcolor=black, filecolor=black, urlcolor=black, citecolor=black }

\newcommand{\coord}{\texttt{<coord>}}
\algrenewcommand\algorithmicrequire{\textbf{Input:}}
\algrenewcommand\algorithmicensure{\textbf{Output:}}

\definecolor{colorfirst}{RGB}{255, 204, 204}
\definecolor{colorsecond}{RGB}{255, 230, 204}
\definecolor{colorthird}{RGB}{255, 251, 214}
\newcommand{\bgfirst}[1]{\colorbox{colorfirst}{\strut\,#1\,}}
\newcommand{\bgsecond}[1]{\colorbox{colorsecond}{\strut\,#1\,}}
\newcommand{\bgthird}[1]{\colorbox{colorthird}{\strut\,#1\,}}
\newcommand{\scorelegend}{%
  Best results in each column are highlighted in \protect\bgfirst{red},
  second-best in \protect\bgsecond{orange},
  and third-best in \protect\bgthird{yellow}.}

\definecolor{hsy}{rgb}{0.51, 0.0, 0.13}

\title{Lens3D: Target-Conditioned Visual Foveation for Fine-Grained 3D Understanding}
\author{
  Junming Huang\textsuperscript{1,2*} \quad
  Shuaiying Hou\textsuperscript{1*} \quad
  Zini Chen\textsuperscript{1*} \\
  Chi Wang\textsuperscript{1} \quad
  Qiang Dai\textsuperscript{2} \quad
  Weiwei Xu\textsuperscript{1} \\[1ex]
  \textsuperscript{1}Zhejiang University \qquad
  \textsuperscript{2}LIGHTSPEED
}
\date{\today}

\begin{document}

\maketitle
{\let\thefootnote\relax\footnotetext{\textsuperscript{*}Equal contribution.}}

\begin{figure}[H]
\centering
\includegraphics[width=\linewidth]{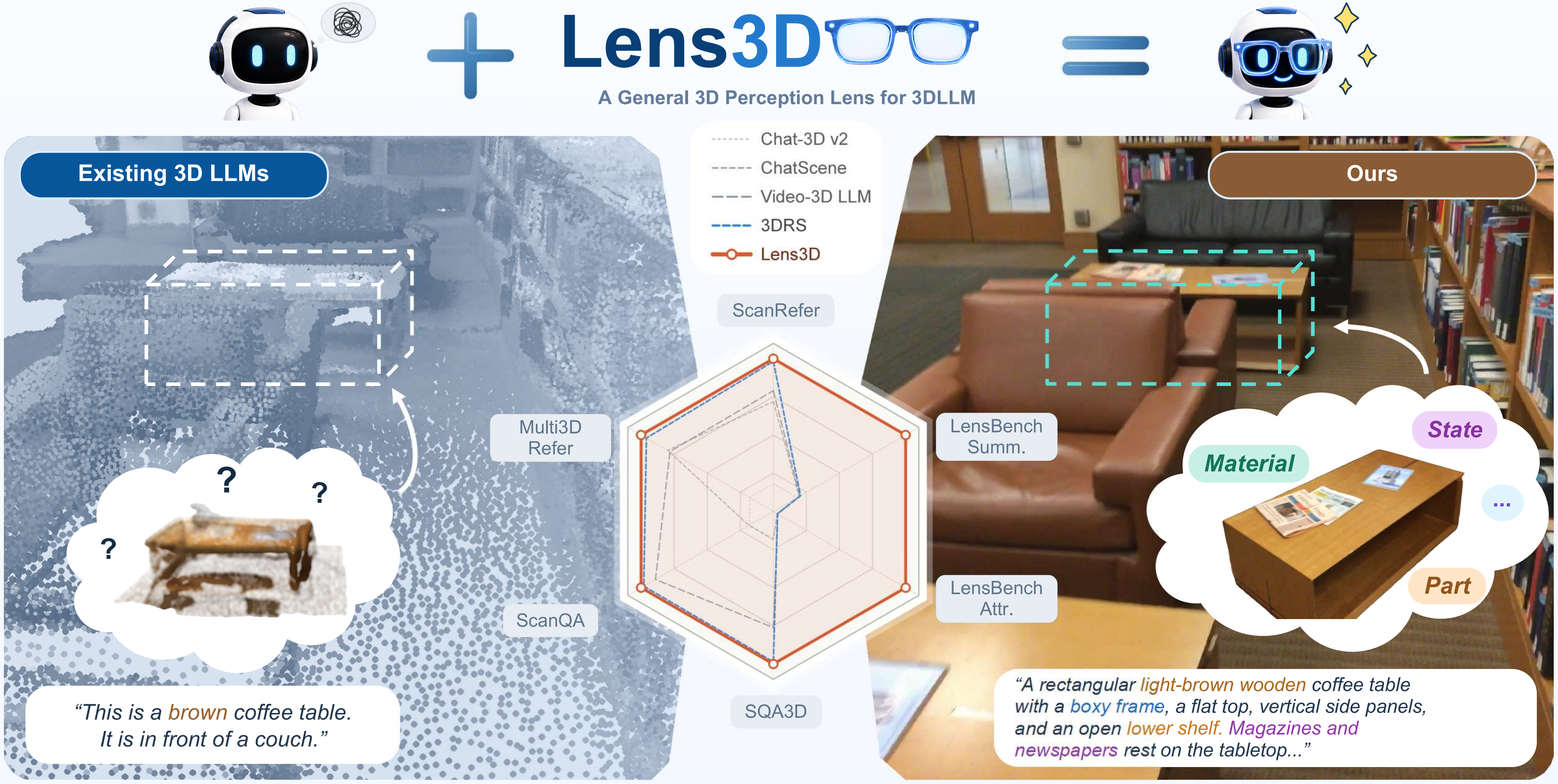}
\caption{\textbf{Lens3D brings fine-grained object understanding to 3D LLMs.}
While existing models capture broad scene structure and salient objects (left), Lens3D further capture instance-level attributes and easily overlooked small objects (right), with the performance radar showing these gains in fine-grained understanding do not come at the cost of the models' original grounding and question-answering abilities (center).
}
\label{fig:teaser}
\end{figure}

\begin{abstract}

Existing 3D large language models often overlook fine-grained attributes and less visually salient objects and parts, even when relevant evidence is present in scene videos. We introduce \emph{Lens3D} to improve fine-grained object understanding through external visual assistance and knowledge transfer. Its \emph{LensUnd} pipeline adopt 3D localization to select informative, complementary views for an external 2D vision-language model, supporting fine-grained object captioning, small-object grounding, and fine-grained object question answering. And its \emph{LensDistill} transfers the resulting fine-grained knowledge to 3D LLMs through detailed caption supervision, enabling captioning from native inputs without external VLM calls. We also construct \emph{LensBench}, a held-out evaluation set of 2,068 objects with three silver-standard reference descriptions per object. Experiments with Video-3D LLM and 3DRS demenstrate that LensDistill substantially improves fine-grained object captioning while preserving existing grounding and scene-level QA performance. These results establish the feasibility of transferring externally acquired fine-grained knowledge into native 3D LLMs.
\end{abstract}

\section{Introduction}



Recent 3D large language models (3D LLMs), such as Video-3D LLM~\citep{zhengVideo3DLLMLearning2025} and 3DRS~\citep{huang3DRSMLLMsNeed2025a}, have advanced 3D scene understanding. However, they often overlook fine-grained object attributes, as well as less visually salient objects and parts (Figure~\ref{fig:teaser}, left). For example, a model may locate an overhead cabinet yet struggle to identify its door material or handle orientation. These details matter for embodied interaction: handle orientation can affect a robot's grasping pose and manipulation strategy. Reliable scene understanding therefore requires both locating objects and recognizing their fine-grained properties.


High-resolution scene videos often contain the necessary visual evidence, but two factors limit its utilization. First, existing 3D datasets emphasize object categories, referring expressions, and short descriptions, providing limited supervision for fine-grained attributes~\citep{dai2017scannet,chen2020scanrefer,chen2021scan2cap}. Second, relevant details may not be visually salient, and their visibility varies across viewpoints. Even when captured in the video, these details may be difficult for a model to distinguish from the surrounding content and use for the task. Together, these challenges motivate more effective utilization of the available visual evidence and richer supervision for fine-grained understanding.

Recent work on 2D vision-language models (VLMs) shows the value of focusing on informative image regions for closer inspection~\citep{zhu2025active,carvalho2026cropvlm}. Inspired by this principle, we exploit 3D localization to guide evidence selection across video frames. Our framework, \emph{Lens3D}, combines externally assisted fine-grained understanding with knowledge transfer to 3D LLMs. Its external understanding pipeline, \emph{LensUnd}, uses a 3D LLM to locate the object or region relevant to a query, such as the cabinet containing the handle of interest. It then selects clear, complementary views from the input video. An external 2D VLM examines these focused evidences to perform fine-grained object captioning, small-object grounding, or fine-grained object question answering (QA). This connects existing 3D grounding capabilities with fine-grained 2D perception.



Beyond external assistance, we ask whether the resulting knowledge can be transferred to the 3D LLM itself. To this end, we introduce \emph{LensDistill}, which uses LensUnd as a teacher through asymmetric privileged-information distillation~\citep{cui2026brief}. We study this transfer through fine-grained object captioning, which provides supervision for materials, parts, states, and spatial relationships. The teacher accesses the selected visual evidence, while the student learns to generate detailed descriptions from its native inputs. After distillation, the student performs captioning without additional view processing or external VLM calls.

Evaluating this capability requires references that capture instance-level details, which existing captioning benchmarks cover sparsely. We therefore construct \emph{LensBench}, a held-out evaluation set of 2,068 objects with three independently generated VLM descriptions per object. A questionnaire-based quality audit supports their use as silver-standard references. Experiments with two representative 3D LLMs, Video-3D LLM and 3DRS, show that LensDistill substantially improves fine-grained object captioning, reaching performance comparable to a similarly sized 2D VLM while preserving existing grounding and scene-level QA performance (Figure~\ref{fig:teaser}, right and center). Qualitative results further illustrate LensUnd's utility for small-object grounding and fine-grained object QA. Our contributions are: 
\begin{itemize}
    \item We introduce \emph{LensUnd}, which uses 3D localization to better exploit visual details already present in scene videos. With an external 2D VLM, it supports fine-grained object captioning, small-object grounding, and fine-grained object QA.
    \item We propose \emph{LensDistill} and demonstrate that detailed caption supervision can transfer externally acquired object knowledge to 3D LLMs. This improves fine-grained object captioning without external VLM calls at inference, while preserving existing grounding and scene-level QA performance.
    \item We construct \emph{LensBench}, covering 2,068 objects with three silver-standard reference descriptions per object. These references support evaluation of instance-level attributes that are sparsely covered by existing 3D captioning benchmarks.
\end{itemize}

\section{Related Work}

\subsection{3D scene understanding}

3D scene understanding spans object detection, semantic and instance segmentation, and language-conditioned tasks such as grounding, captioning, and question answering. ScanRefer~\citep{chen2020scanrefer}, ReferIt3D~\citep{achlioptas2020referit3d}, and Multi3DRefer~\citep{zhang2023multi3drefer} study 3D target localization from natural-language expressions, while 3DVG-Transformer~\citep{zhao20213dvg}, BUTD-DETR~\citep{jain2022bottom}, and EDA~\citep{wu2023eda} improve grounding through cross-modal scene modeling. Scan2Cap~\citep{chen2021scan2cap} introduced 3D dense captioning; D3Net~\citep{chen2022d} and Vote2Cap-DETR~\citep{chen2023end} subsequently advanced joint grounding and end-to-end captioning, respectively, and PQ3D~\citep{zhu2024unifying} uses promptable queries to unify multiple 3D vision--language tasks. ScanQA~\citep{azuma2022scanqa} and SQA3D~\citep{ma2022sqa3d} further evaluate scene-level spatial reasoning. These tasks and datasets, however, primarily emphasize object categories, and spatial relations, with limited coverage of fine-grained object details.

\subsection{3D large language models}

3D large language models connect 3D scene representations with pretrained language models to support grounding, description, question answering, and even generation within a unified framework. For 3D understanding, 3D-LLM~\citep{hong20233d}, Chat-3D~\citep{wang2023chat}, LL3DA~\citep{chen2024ll3da}, and LEO~\citep{huang2023embodied} explore different interfaces between point clouds, object tokens, and language models, while LLaVA-3D~\citep{zhu2024llava} augments multiview 2D features with 3D spatial information. Beyond understanding, ShapeLLM-Omni~\citep{ye2025shapellm} and CG-MLLM~\citep{huang2026cgmllmcaptioninggenerating3d} extend 3D multimodal models toward unified understanding and generation, enabling them to both reason about and synthesize 3D content. 
However, incorporating 3D perception into pretrained language models typically requires additional 3D encoders or alignment modules, limiting their ability to directly inherit the rich knowledge of large-scale pretrained models. Video-based approaches provide an alternative by representing multiview 3D observations as video, allowing them to directly exploit the strong visual and temporal understanding capabilities of pretrained models.Video-3D LLM~\citep{zhengVideo3DLLMLearning2025} treats multiview indoor observations as video and injects 3D positional encodings, while 3DRS~\citep{huang3DRSMLLMsNeed2025a} further introduces explicit 3D representation supervision. Despite these increasingly capable architectures and their access to rich multimodal observations, existing 3D LLMs remain constrained by datasets that emphasize categories, spatial relations, and other coarse supervision~\citep{dai2017scannet,chen2020scanrefer,chen2021scan2cap}. Consequently, their understanding remains coarse, with limited training signal for fine-grained object details and small, easily overlooked objects.

\subsection{Distillation with privileged information}

Knowledge distillation~\citep{hinton2015distilling} transfers a teacher's outputs or representations to a student, while Learning Using Privileged Information~\citep{vapnik2009new} allows the teacher to access additional observations that are unavailable to the deployed student. Generalized distillation~\citep{lopez2015unifying} combines these ideas under asymmetric teacher--student observations. In 3D understanding, SAT~\citep{yang2021sat} uses training-time 2D semantics to improve point-cloud grounding. X-Trans2Cap~\citep{yuan2022x} transfers 2D appearance knowledge from a multimodal teacher to a point-cloud captioner, while ExCap3D~\citep{yeshwanth2025excap3d} uses visibility-ranked multiview images and a VLM to construct fine-grained descriptions for training a dedicated captioning model. Vision-OPD~\citep{yuan2026vision} similarly treats target-region crops as privileged visual input and transfers regional perception to the full-image policy through on-policy self-distillation. Lens3D extends this direction with a target-conditioned evidence interface shared across fine-grained object captioning, small-object grounding, and fine-grained object QA. LensUnd selects views by marginal target-surface coverage. LensDistill then uses captions generated from these views to train general 3D LLMs. Supplementary Sections~\ref{supp:view-coverage} and~\ref{supp:attribute-analysis} compare view coverage and description quality across different selection strategies.

\begin{figure*}[t]
\centering
\includegraphics[width=0.9\textwidth]{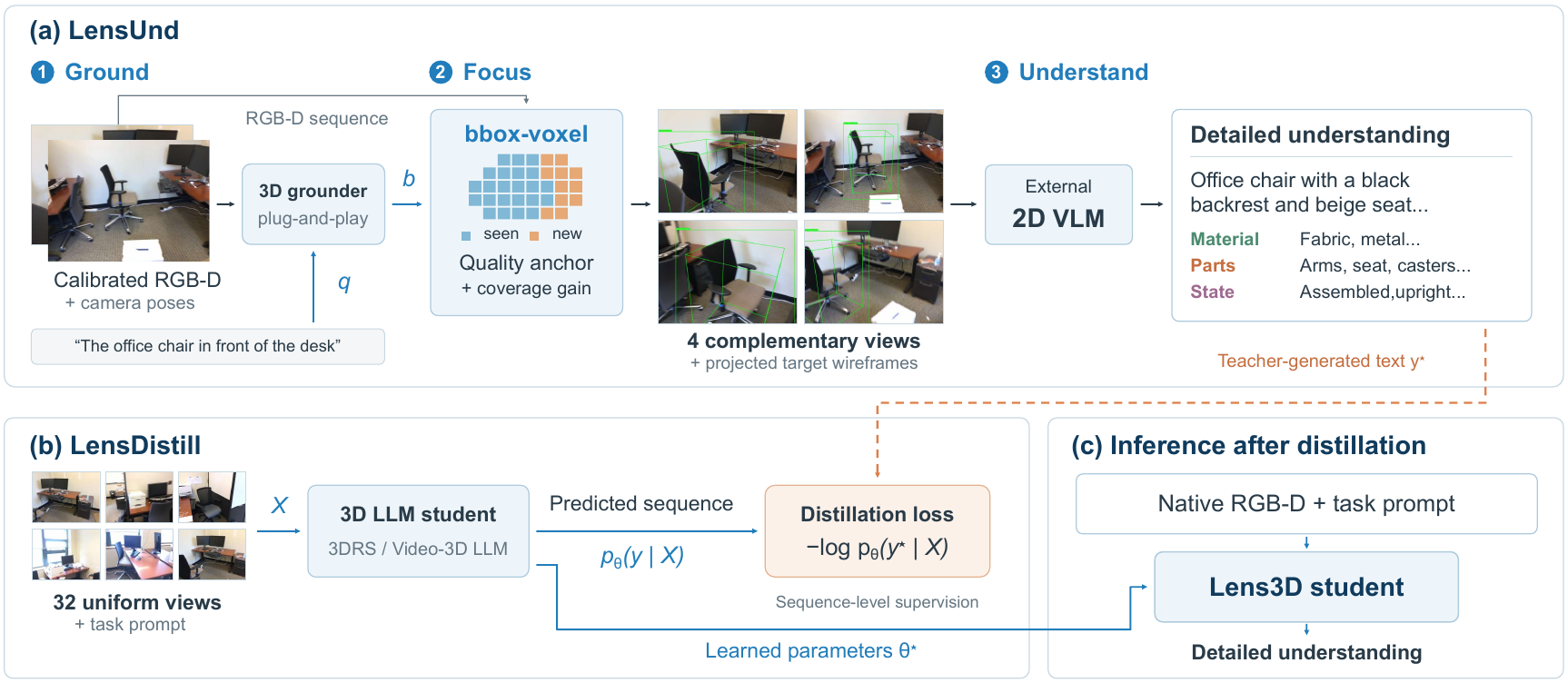}
\caption{\textbf{Overview of Lens3D.} (a) LensUnd follows Ground--Focus--Understand: a 3D grounder localizes an anchor box $b$ from a language query $q$, bbox-voxel selects complementary views by coverage gain and projects the target wireframe onto them, and an external 2D VLM produces a detailed understanding $y^\star$ of the target. (b) LensDistill trains a 3D LLM student with the sequence-level distillation loss $-\log p_\theta(y^\star \mid X,c(b),p)$, using teacher-generated $y^\star$ as privileged supervision while the student receives uniformly sampled native views, the target center $c(b)$ encoded as \coord, and a task prompt. (c) After distillation, the external LensUnd path is removed; the student produces the same detailed understanding from native RGB-D, the target-center, and a task prompt.}
\label{fig:method}
\end{figure*}

\section{Method}

Lens3D consists of LensUnd, a general target-conditioned understanding pipeline, and LensDistill, an asymmetric privileged-information distillation method. As illustrated in Figure~\ref{fig:method}, LensUnd follows a Ground--Focus--Understand scheme. It first uses a 3D LLM to locate the object or region relevant to a language query. The resulting 3D bounding box guides the selection of complementary RGB views, in which the target region is explicitly marked. An external 2D VLM then processes these views according to a task-specific instruction. Through different instructions and output formats, the same evidence collection procedure supports fine-grained object captioning, small-object grounding, and fine-grained object QA. LensDistill then transfers the resulting fine-grained knowledge to a 3D LLM operating on its native inputs. We study this transfer through fine-grained object captioning, using structured, detailed captions as supervision. After distillation, the student generates detailed descriptions without additional view processing or external VLM calls at inference.

\subsection{Problem setting}

Let an indoor scene be
\begin{equation}
S=\{(I_t,D_t,\Pi_t)\}_{t=1}^{T},
\end{equation}
where $I_t$, $D_t$, and $\Pi_t$ denote the RGB image, depth map, and camera parameters of frame $t$. 
Given a language query $q$, a 3D grounder $G$ predicts an bounding box $b$ for the relevant object or region. LensUnd constructs target-conditioned visual evidence $\mathcal{E}_b$, which an external 2D VLM $U$ processes under task instruction $p$.
LensDistill trains a student $f_\theta$ using its native scene input $X$, the target center $c(b)$ encoded by the \coord\ token, and instruction $p$:
\begin{equation}
b=G(S,q),\qquad y^\star=U(\mathcal{E}_b,p),\qquad y=f_\theta(X,c(b),p).
\end{equation}
Depth and camera parameters support target visibility estimation, 3D-box projection, and view selection. Following Video-3D LLM and 3DRS, $X$ comprises subsampled RGB views with depth- and camera-derived 3D positional encodings for their visual tokens. We perform distillation through fine-grained object captioning, using the external VLM's descriptions as teacher responses $y^\star$. LensUnd also supports small-object grounding and fine-grained object QA by varying $p$.

\subsection{Target-conditioned evidence acquisition}


Given target box $b$, RGB-D sequence $S$, and frame budget $K$, the Focus stage selects complementary views that clearly show the target region and explicitly marks that region in each image.

\paragraph{Frame quality and target-surface voxels.}

Let $\mathcal{C}$ contain geometrically valid, depth-visible candidate frames. We assign each candidate a quality score $s_t$ based on projected target completeness, image coverage, centrality, and occlusion estimated through depth consistency. We then unproject its depth pixels, retain 3D points inside $b$, and discretize them into an occupied voxel set $\mathcal{V}_t$. All views use a shared voxel grid with approximately $n_v=12$ cells along the shortest box axis. Thus, $\mathcal{V}_t$ represents observed surfaces within the target region rather than its projected image area.


\paragraph{Greedy view selection.}
Selecting frames solely by quality can yield redundant views. We instead initialize the selected set $\mathcal{S}$ with the highest-quality frame, which has the highest $s_t$, and greedily maximize additional surface coverage:
\begin{equation}
\Delta(t\mid\mathcal{S})
=
\left|
\mathcal{V}_t
\setminus
\bigcup_{t'\in\mathcal{S}}\mathcal{V}_{t'}
\right|.
\end{equation}
Selected frames must be separated by at least $\delta$ frames. When no feasible candidate adds new voxels, we fill the remaining slots by quality under the same temporal constraint. Selection ends when $K$ frames are chosen or no feasible candidates remain. We use $K=4$ and $\delta=12$ by default. We call this bounding-box voxel-selection strategy \emph{bbox-voxel}; Algorithm~\ref{alg:bbox-voxel} gives the complete procedure.




\begin{algorithm}[t]
\caption{BBox-voxel frame selection}
\label{alg:bbox-voxel}
\small
\begin{algorithmic}[1]
\Require Scene $S$, target box $b$, frame budget $K$, temporal gap $\delta$, voxel resolution $n_v$
\Ensure Selected frame indices $\mathcal{S}$, where $|\mathcal{S}|\leq K$
\State Construct candidate set $\mathcal{C}$ from $S$ and $b$
\ForAll{$t\in\mathcal{C}$}
    \State Compute frame quality $s_t$ and visible voxels $\mathcal{V}_t$
\EndFor
\State $\mathcal{S}\gets\varnothing$; $\mathcal{U}\gets\varnothing$
\While{$|\mathcal{S}|<K$}
    \State $\mathcal{C}'\gets\mathcal{C}\setminus\mathcal{S}$
    \State Remove $t\in\mathcal{C}'$ if $|t-t'|<\delta$ for any $t'\in\mathcal{S}$
    \If{$\mathcal{C}'=\varnothing$}
        \State \textbf{break}
    \EndIf
    \ForAll{$t\in\mathcal{C}'$}
        \State $g_t\gets|\mathcal{V}_t\setminus\mathcal{U}|$
    \EndFor
    \If{$\mathcal{S}=\varnothing$ \textbf{or} $\max_{t\in\mathcal{C}'}g_t=0$}
        \State $t^\star\gets\arg\max_{t\in\mathcal{C}'}s_t$
    \Else
        \State $t^\star\gets\arg\max_{t\in\mathcal{C}'}g_t$
    \EndIf
    \State $\mathcal{S}\gets\mathcal{S}\cup\{t^\star\}$
    \State $\mathcal{U}\gets\mathcal{U}\cup\mathcal{V}_{t^\star}$
\EndWhile
\State \Return $\mathcal{S}$
\end{algorithmic}
\end{algorithm}


\paragraph{Target marking.}
We project the twelve edges of $b$ onto each selected RGB image using its camera parameters, obtaining the resulting wireframe $\widetilde I_t$ and evidence $\mathcal{E}_b$:
\begin{equation}
\widetilde I_t=\operatorname{Highlight}(I_t,b,\Pi_t),
\qquad
\mathcal{E}_b=\{\widetilde I_t\}_{t\in\mathcal{S}}.
\end{equation}
This operation supplies an explicit instance-level spatial condition to the 2D VLM, allowing it to identify the object to describe when multiple same-class objects are visible.

\subsection{LensUnd: A plug-and-play understanding pipeline}


LensUnd follows Ground--Focus--Understand pipeline as following:
\begin{equation}
b=G(S,q),\qquad
\{\widetilde I_{t_k}\}_{k=1}^{K}=\operatorname{Focus}(S,b),\qquad
r=U(\{\widetilde I_{t_k}\}_{k=1}^{K},p),
\end{equation}
where $G$ is a 3D grounding model, $\operatorname{Focus}$ is the target-conditioned evidence process in Section~3.2, $U$ is a 2D VLM, $p$ is a downstream-task instruction, and $r$ is the task-specific response. In the standard setting, $q$ is a ScanRefer-style referring expression~\citep{chen2020scanrefer}.
LensUnd requires only the predicted box from $G$ and does not
modify the grounder's architecture.

\begin{figure*}[t]
\centering
\includegraphics[width=0.95\textwidth]{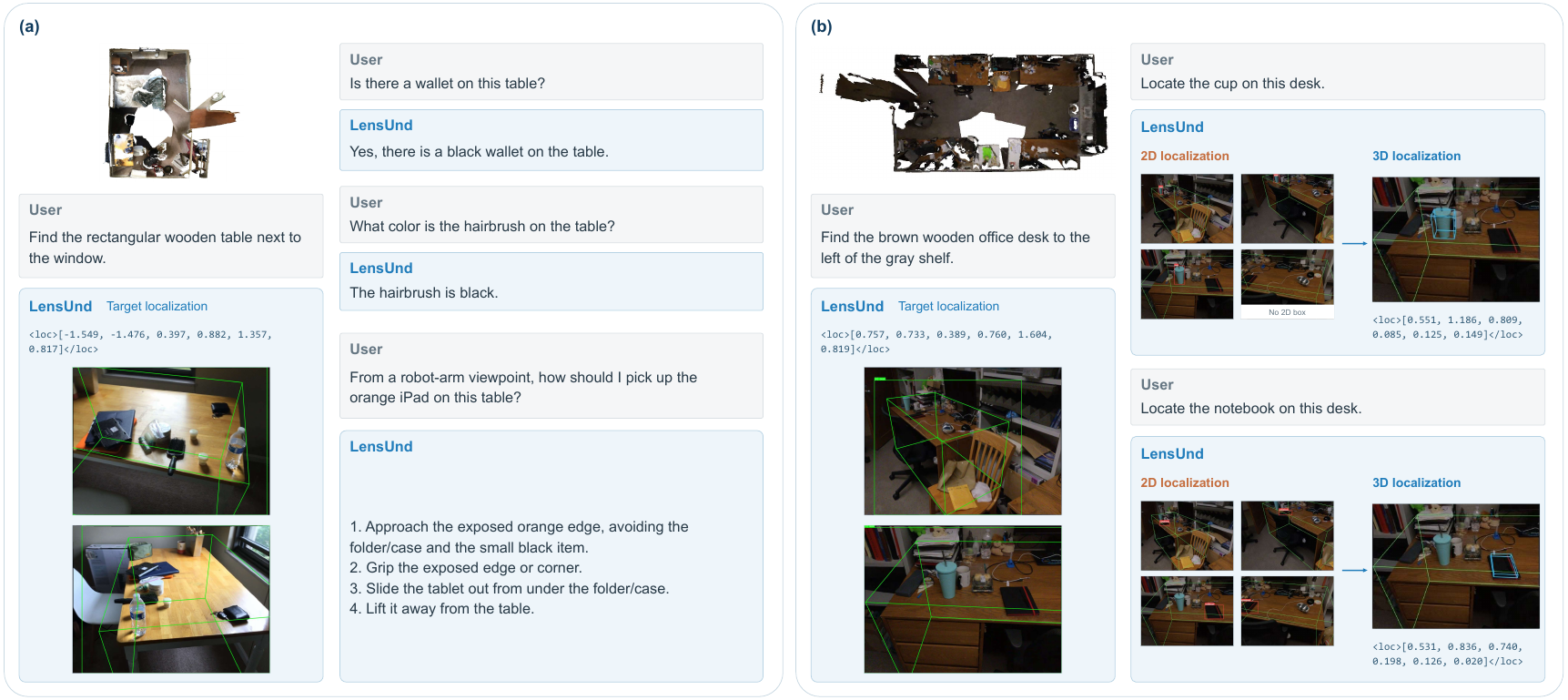}
\caption{
\textbf{Fine-grained interactions with LensUnd.}
Complementary views selected by LensUnd support (a) object QA on presence, appearance, and manipulation, and (b) small-object grounding by combining per-view 2D predictions into a 3D box using enclosed depth and camera parameters.
}

\label{fig:lensund-tasks}
\end{figure*}

The same evidence selection procedure supports fine-grained object captioning, small-object grounding, and fine-grained object QA through different instructions and output formats. For small-object grounding, the initially grounded region can contain the smaller object of interest. LensUnd therefore accommodates compatible 3D grounders and 2D VLMs without changing the evidence selection procedure. We quantitatively evaluate fine-grained object captioning with Video-3D LLM and 3DRS; Figure~\ref{fig:lensund-tasks} illustrates the other two tasks.


\subsection{LensDistill: Asymmetric distillation with privileged information}

LensUnd relies on the full RGB-D sequence, target-conditioned view selection, and an additional 2D VLM call, introducing inference cost when directly deploying. To avoid this cost, we propose LensDistill, an asymmetric teacher--student framework for transferring externally acquired fine-grained knowledge into native 3D LLMs. During training, the teacher accesses privileged target-level evidence, while the student operates on the same observation space used at test time. At inference, the student produces the corresponding fine-grained understanding without privileged evidence or external VLM calls. We perform and validate this transfer paradigm on fine-grained object captioning.


\paragraph{Teacher and student inputs.}
For a target region represented by a 3D box $b$, LensUnd first constructs privileged evidence $\mathcal{E}_b=\{\widetilde I_t\}_{t\in\mathcal{S}}$, from which an external VLM teacher generates a fine-grained understanding $y^\star$ under task instruction $p$. The student retains the base 3D LLM's native scene input $X$, the target center $c(b)$ encoded by the \coord\ token, and instruction $p$. The selected, wireframe-marked views are used only to generate teacher supervision. LensDistill thus transfers knowledge acquired from privileged target-level observations rather than matching teacher and student responses under identical inputs.

\paragraph{Teacher supervision and training.}
For training target region specified by $b$, LensUnd constructs target-conditioned evidence and invokes the chosen 2D VLM to generate $y^\star$ as supervision for observation $X$. This formulation allows supervision to be extended to new scenes, objects, instructions, or teacher models. Given $y^\star=(y_1^\star,\ldots,y_L^\star)$, we train the student with the autoregressive objective
\begin{equation}
\mathcal{L}_{\mathrm{distill}}
=-\sum_i\log p_\theta(y_i^\star\mid y_{<i}^\star,X,c(b),p)
\end{equation}
where $p_\theta$ is the student's conditional token distribution. After distillation, the student generates detailed descriptions directly from $(X,c(b),p)$ without additional view processing or 2D VLM calls.

\section{Experiments}

LensUnd demonstrates small-object grounding and fine-grained object QA over focused multiview evidence. Fine-grained object captioning provides the quantitative testbed for establishing the benefit of target-conditioned evidence and LensDistill’s transfer effectiveness across different 3D LLMs.

\subsection{Quantitative evaluation}

\paragraph{LensBench construction.}
We construct LensBench to evaluate fine-grained target descriptions against diverse, independent references. It contains 2,068 validation objects from 141 ScanNet validation scenes, with three independent references generated by GPT 5.4~\citep{singh2025openai}, Gemini 3~\citep{team2023gemini}, and Qwen 3.5~\citep{team2026qwen3} from views focused using ground-truth boxes. Each reference contains a summary and structured fields for visible parts, materials, and object state; Parts, Materials, and State F1 measure set-token overlap on the corresponding fields. Its scenes do not overlap the 561 ScanNet training scenes used by LensDistill. In a public questionnaire, respondents evaluate 100 randomly sampled objects and separately rate each anonymized reference against the focused views, yielding 300 ratings with a mean of 3.85/4.0 and supporting their use as high-quality silver references.

\begin{table}[t]
\centering
\caption{Core LensBench results. ``Attr.\ Mean'' averages Parts, Materials, and State F1 and is reported only for models that produce the required structured fields. Human satisfaction (Human Sat.dcfr) is rated on a four-point scale. \scorelegend}
\label{tab:lensbench}
\resizebox{0.72\linewidth}{!}{%
\begin{tabular}{lrrrr}
\toprule
Method & Summary F1 & Summary Sem. & Attr.\ Mean & Human Sat. \\
\midrule
Qwen3-VL-2B~\citep{bai2025qwen3} & 0.380 & 0.686 & 0.356 & -- \\
Qwen3-VL-2B (LensUnd) & 0.445 & \cellcolor{colorthird}0.747 & 0.397 & -- \\
Qwen3-VL-8B (LensUnd) & \cellcolor{colorsecond}0.485 & \cellcolor{colorfirst}\textbf{0.783} & \cellcolor{colorthird}0.528 & -- \\
\midrule
Chat-3D v2 & 0.222 & 0.522 & -- & -- \\
ChatScene & 0.227 & 0.546 & -- & -- \\
Video-3D LLM & 0.228 & 0.542 & -- & 1.88 \\
\textbf{Lens3D (Video-3D LLM)} & \cellcolor{colorthird}0.483 & \cellcolor{colorthird}0.747 & \cellcolor{colorsecond}0.533 & 3.71 \\
3DRS & 0.230 & 0.549 & -- & 1.92 \\
\textbf{Lens3D (3DRS)} & \cellcolor{colorfirst}\textbf{0.489} & \cellcolor{colorsecond}0.752 & \cellcolor{colorfirst}\textbf{0.538} & \textbf{3.86} \\

\bottomrule
\end{tabular}
}
\end{table}

\paragraph{Benchmarks and metrics.}
We evaluate fine-grained descriptions on LensBench and retained native capabilities on the standard ScanNet~\citep{dai2017scannet} benchmarks ScanRefer~\citep{chen2020scanrefer}, Multi3DRefer~\citep{zhang2023multi3drefer}, ScanQA~\citep{azuma2022scanqa}, and SQA3D~\citep{ma2022sqa3d}. We report Acc@0.25/0.5 for ScanRefer, F1@0.25/0.5 for Multi3DRefer, CIDEr~\citep{vedantam2015cider} for ScanQA, and exact match for SQA3D. Description metrics include token F1 and embedding similarity for summaries and set-token F1 for parts, materials, and state.

\paragraph{Implementation details.}
We instantiate students with Video-3D LLM~\citep{zhengVideo3DLLMLearning2025} and 3DRS~\citep{huang3DRSMLLMsNeed2025a}. For 7,854 objects, LensUnd provides four target-focused views independently to GPT 5.4, Gemini 3, and Qwen 3.5, chosen for their strong general-purpose visual reasoning and complementary model priors; all successful responses are retained, yielding 23,560 structured captions. We jointly fine-tune each student's main vision--language stack for one epoch, replacing short Scan2Cap descriptions with this supervision while retaining the original grounding and QA data. Students keep their native 32-view inputs. Supplementary Section~\ref{supp:reproducibility} gives the complete prompt, data composition, frame-quality definition, and optimization details.

\paragraph{Evaluation of fine-grained object understanding.}
The quantitative results in Table~\ref{tab:lensbench} show consistent fine-grained description gains for both student architectures. LensUnd also benefits from target-conditioned evidence and stronger VLM capacity, while remaining robust to predicted target boxes. In a blind user study on 100 objects using a four-point scale, LensDistill raises the mean rating from 1.92 to 3.86 for 3DRS and from 1.88 to 3.71 for Video-3D LLM, confirming substantial improvements in perceived description quality.

\paragraph{Ablations study.}
Supplementary Table~\ref{tab:lensbench-full} reports the complete LensBench results. Wireframe marking contributes to performance, increasing Summary F1 from 0.263 to 0.370 with our selected views. With ground-truth boxes, four views, projected wireframes, and the same VLM, our bbox-voxel frame selection achieves a Summary F1 of 0.445, outperforming ExCap3D's visibility ranking (0.436) and Video-3D LLM's greedy selection used in object-level (0.353). Increasing temporal uniform sampling from 4 to 32 views yields only marginal gains, possibly since the additional views provide little target-relevant evidence. Using predicted boxes retains at least 97.5\% of the performance obtained with ground-truth boxes across the main metrics. Scaling the VLM from 2B to 8B further improves Summary F1 from 0.445 to 0.485 and Attr.\ Mean from 0.397 to 0.528. The corresponding gains for parts, materials, and states are 22.2, 9.4, and 7.6 percentage points, respectively.

\paragraph{Retained 3D capabilities.}
Table~\ref{tab:standard_3d_benchmarks} shows that, under its detailed-captioning instantiation, LensDistill enhances the fine-grained description capability that is limited in the base models while preserving, and in some cases improving, their original grounding and question-answering performance. The gains are most apparent in grounding. Table~\ref{tab:unique-multiple} further shows that both backbones improve at both thresholds when same-class distractors are present, whereas unique-instance scores change by at most 0.54 points. This concentration of gains in challenging multi-instance scenes suggests that detailed-caption supervision improves instance discrimination while retaining standard grounding performance.

\begin{table*}[t]
  \centering
  \caption{Comparison on standard 3D scene-understanding benchmarks. Asterisks indicate results reproduced in our local environment and evaluation stack. Lens3D uses only the corresponding base model's native input at inference. \scorelegend}
  \label{tab:standard_3d_benchmarks}
  \scriptsize
  \resizebox{\textwidth}{!}{%
  \begin{tabular}{lcccccc}
  \toprule
  & \multicolumn{2}{c}{ScanRefer} & \multicolumn{2}{c}{Multi3DRefer} &
  ScanQA & SQA3D \\
  \cmidrule(lr){2-3}\cmidrule(lr){4-5}\cmidrule(lr){6-6}\cmidrule(lr){7-7}
  Method & Acc@0.25 & Acc@0.5 & F1@0.25 & F1@0.5 & CIDEr & EM \\
  \midrule
  \multicolumn{7}{l}{\textit{Task-specific models}} \\
  ScanRefer~\citep{chen2020scanrefer} & 37.3 & 24.3 & -- & -- & -- & -- \\
  MVT~\citep{huang2022multi} & 40.8 & 33.3 & -- & -- & -- & -- \\
  3DVG-Transformer~\citep{zhao20213dvg} & 45.9 & 34.5 & -- & -- & -- & -- \\
  ViL3DRel~ \citep{chen2022language} & 47.9 & 37.7 & -- & -- & -- & -- \\
  M3DRef-CLIP~\citep{zhang2023multi3drefer} & 51.9 & 44.7 & 42.8 & -- & -- & -- \\
  ScanQA~\citep{azuma2022scanqa} & -- & -- & -- & -- & 64.9 & 47.2 \\
  3D-VisTA~\citep{zhu20233d} & 50.6 & 45.8 & -- & -- & 69.6 & 48.5 \\
  \midrule
  \multicolumn{7}{l}{\textit{Generalist models}} \\
  3D-LLM (Flamingo)~\citep{hong20233d} & 21.2 & -- & -- & -- & 59.2 & -- \\
  3D-LLM (BLIP2-FlanT5)~\citep{hong20233d} & 30.3 & -- & -- & -- & 69.4 & -- \\
  Chat-3D~\citep{wang2023chat} & -- & -- & -- & -- & 53.2 & -- \\
  Chat-3D v2~\citep{huang2023chat} & 42.5 & 38.4 & 45.1 & 41.6 & 87.6 & 54.7 \\
  LL3DA~\citep{chen2024ll3da} & -- & -- & -- & -- & 76.8 & -- \\
  Scene-LLM~\citep{fu2024scene} & -- & -- & -- & -- & 80.0 & 53.6 \\
  LEO~\citep{huang2023embodied} & -- & -- & -- & -- & 101.4 & 50.0 \\
  Grounded 3D-LLM~\citep{chen2024grounded} & 47.9 & 44.1 & 45.2 & 40.6 & 72.7 & -- \\
  PQ3D~\citep{zhu2024unifying} & 57.0 & 51.2 & -- & 50.1 & -- & 47.1 \\
  ChatScene~\citep{huang2024chat} & 55.5 & 50.2 & 57.1 & 52.4 & 87.7 & 54.6 \\
  LLaVA-3D~\citep{zhu2024llava} & 54.1 & 42.4 & -- & -- & 91.7 & 55.6 \\
  Inst3D-LMM~\citep{yu2025inst3d} & 57.8 & 51.6 & 58.3 & 53.5 & 88.6 & -- \\
  3D-LLaVA~\citep{deng20253d} & 51.2 & 40.6 & -- & -- & 92.6 & 54.5 \\
  Video-3D LLM*~\citep{zhengVideo3DLLMLearning2025} & 58.2 & 51.8 & 57.4 & 52.1 & \cellcolor{colorthird}102.1 & \cellcolor{colorthird}58.5 \\
  \textbf{Lens3D (Video-3D LLM)} & \cellcolor{colorthird}\textbf{60.3} & \cellcolor{colorthird}\textbf{53.7} & \cellcolor{colorthird}\textbf{58.9} & \cellcolor{colorthird}\textbf{53.6} & 101.2 & 58.2 \\
  3DRS*~\citep{huang3DRSMLLMsNeed2025a} & \cellcolor{colorsecond}62.7 & \cellcolor{colorsecond}56.0 & \cellcolor{colorsecond}60.6 & \cellcolor{colorsecond}55.2 & \cellcolor{colorsecond}103.9 & \cellcolor{colorsecond}60.0 \\
  \textbf{Lens3D (3DRS)} & \cellcolor{colorfirst}\textbf{63.1} & \cellcolor{colorfirst}\textbf{56.4} & \cellcolor{colorfirst}\textbf{61.3} & \cellcolor{colorfirst}\textbf{55.9} & \cellcolor{colorfirst}\textbf{104.3} & \cellcolor{colorfirst}\textbf{60.1} \\
  \bottomrule
  \end{tabular}}
  \end{table*}

\begin{table}[t]
\centering
\caption{ScanRefer~\citep{chen2020scanrefer} grounding on unique and multiple subsets.}
\label{tab:unique-multiple}
\resizebox{0.5\textwidth}{!}{%

\begin{tabular}{lrrrr}
\toprule
Method & \multicolumn{2}{c}{Unique} & \multicolumn{2}{c}{Multiple} \\
\cmidrule(lr){2-3}\cmidrule(lr){4-5}
& @0.25 & @0.5 & @0.25 & @0.5 \\
\midrule
Video-3D LLM & 87.80 & 78.16 & 51.06 & 45.40 \\
Lens3D (Video-3D LLM) & 87.26 & 78.16 & 53.84 & 47.87 \\
3DRS & 86.72 & 77.40 & 56.87 & 50.79 \\
Lens3D (3DRS) & 86.45 & 77.56 & 57.44 & 51.25 \\
\bottomrule
\end{tabular}
}
\end{table}

\subsection{Qualitative evaluation}

\begin{figure*}[t]
\centering
\includegraphics[width=0.95\textwidth]{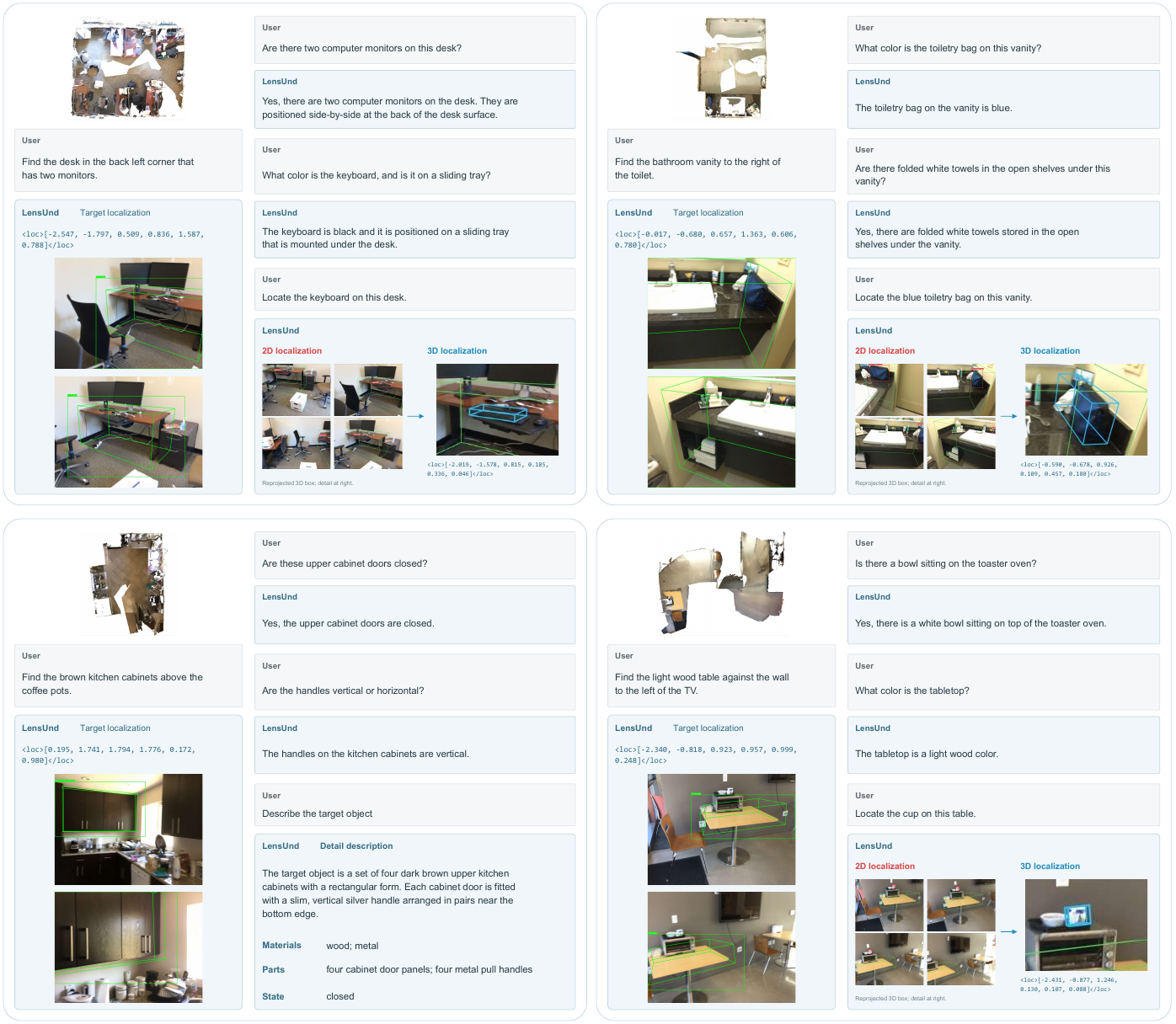}
\vspace{-0.5em}
\caption{\textbf{More qualitative results with LensUnd.} Examples across multiple scenes demonstrate LensUnd’s ability to handle varied fine-grained 3D understanding scenarios.}
\label{fig:qualitative-evaluation}
\end{figure*}

Beyond fine-grained object captioning, LensUnd supports small-object grounding and fine-grained object QA through task-specific instructions over focused multiview evidence. For small-object grounding, it focuses on a coarse anchor region, localizes a queried object in the selected views, and aggregates the unprojected observations into a 3D box. For fine-grained object QA, the external VLM reasons over the focused evidence to answer questions about object presence, appearance, and manipulation. Figures~\ref{fig:lensund-tasks} and~\ref{fig:qualitative-evaluation} present representative examples of grounding and QA.

\subsection{Evidence analysis}

We use two complementary tests to establish that LensDistill uses target-level visual evidence. Masking the target region causes a substantial performance drop: Summary F1 decreases from 0.489 to 0.369, Overall F1 from 0.470 to 0.337, and Summary Sem.\ from 0.753 to 0.560. In the final four layers, the top 5\% of image patches receive 35.0\% of answer-token attention on average, seven times the uniform expectation, while attention alignment with the projected target reaches 8.96 times the uniform expectation. Together, these results show that LensDistill retrieves target appearance from native multiview observations rather than relying only on language patterns or coordinate priors. Supplementary Section~\ref{supp:visual-analysis} provides the masking results, full protocol, and layer-wise analysis.

\section{Conclusion}

We presented Lens3D, a framework that combines target-conditioned visual foveation with capability transfer for fine-grained 3D understanding. LensUnd constructs focused multiview evidence that enables an external VLM to supplement existing 3D LLMs, while LensDistill transfers the acquired knowledge back into 3D LLMs operating on their native inputs. Experiments with two representative 3D LLMs show that caption-based LensDistill substantially improves fine-grained object descriptions while maintaining their native grounding and question-answering capabilities.


\clearpage
\printbibliography

\clearpage
\appendix
\stepcounter{section}
\hypertarget{appendix.\thesection}{}

\subsection{Experimental details}
\label{supp:experimental-details}

\paragraph{Datasets and tasks.}
We evaluate existing grounding and question-answering capabilities on four standard 3D scene-understanding benchmarks. ScanRefer~\citep{chen2020scanrefer} evaluates single-target 3D referring-expression grounding. Multi3DRefer~\citep{zhang2023multi3drefer} extends grounding to zero-, single-, and multiple-target expressions. ScanQA~\citep{azuma2022scanqa} and SQA3D~\citep{ma2022sqa3d} evaluate scene question answering and situated spatial reasoning, respectively. All are based on ScanNet~\citep{dai2017scannet} scenes, and we follow the standard splits used by prior work. Our local evaluations contain 9,508 ScanRefer examples, 11,120 Multi3DRefer examples, 4,675 ScanQA questions, and 3,519 SQA3D questions.

For object description, we replace the short-text-centered Scan2Cap~\citep{chen2021scan2cap} evaluation with LensBench. LensBench contains 2,068 targets, each with three independently generated detailed references from GPT, Qwen, and Gemini. The references cover an overall summary, materials, parts, state, visible details, and spatial relations. LensBench is used only for evaluation and is strictly separated from LensDistill training samples.

\paragraph{Human reference audit.}
We publicly distribute a questionnaire that samples 100 LensBench objects uniformly at random without replacement. Each question shows the four target-focused views and the three model identities are hidden as ``Data 1--3.'' Respondents score each reference separately using four anchors: 1, incorrect (wrong target or mostly incorrect key facts); 2, partially correct (some visible facts but clear hallucinations or missing key evidence); 3, mostly correct (the target and main details are correct, with only minor inaccuracies); and 4, fully correct (consistent with the images and no evident hallucination). Each reference receives one independent rating, yielding a mean of 3.85/4.0 across 300 judgments.

\paragraph{Metrics.}
For ScanRefer~\citep{chen2020scanrefer}, we report 3D-box grounding accuracy at IoU thresholds 0.25 and 0.5 (Acc@0.25/0.5). Multi3DRefer~\citep{zhang2023multi3drefer} uses F1 at the same thresholds. ScanQA~\citep{azuma2022scanqa} reports CIDEr~\citep{vedantam2015cider}, and SQA3D~\citep{ma2022sqa3d} reports exact match (EM). On LensBench, each prediction is compared with all three references and the scores are averaged. Our principal metrics are token F1 for the target summary (Summary F1) and semantic similarity (Summary Sem.), where the latter is cosine similarity between all-MiniLM-L6-v2 sentence embeddings~\citep{reimers2019sentence}. Summary F1 measures lexical agreement, whereas semantic similarity is more robust to synonymous paraphrases.

\paragraph{Base models and training.}
We use Video-3D LLM~\citep{zhengVideo3DLLMLearning2025} and 3DRS~\citep{huang3DRSMLLMsNeed2025a} as two independent base models. Both construct 3D-aware visual representations from calibrated multiview RGB-D observations and report results on the same indoor-scene tasks. The student uniformly samples 32 views per scene; LensUnd selects four target views with bbox-voxel. An external VLM teacher generates LensDistill supervision from the target-conditioned LensUnd evidence under a detailed-description instruction. Training retains each base model's ScanRefer~\citep{chen2020scanrefer}, Multi3DRefer~\citep{zhang2023multi3drefer}, ScanQA~\citep{azuma2022scanqa}, and SQA3D~\citep{ma2022sqa3d} data and replaces its short Scan2Cap~\citep{chen2021scan2cap} descriptions with detailed supervision.

\subsection{LensDistill reproducibility details}
\label{supp:reproducibility}

\paragraph{Teacher supervision and split.}
We start from 7,875 targets in the ScanRefer training split and successfully annotate 7,854 objects from 561 ScanNet training scenes. For every target, bbox-voxel selects four 1296$\times$968 RGB frames using its ground-truth box and overlays the projected 3D wireframe. GPT 5.4~\citep{singh2025openai}, Gemini 3~\citep{team2023gemini}, and Qwen 3.5~\citep{team2026qwen3} are queried independently with the same evidence and provider-default decoding settings. We retain each successful response as a separate training example rather than merging teacher outputs, producing 23,560 captions. LensBench contains 6,204 references for 2,068 objects from 141 ScanNet validation scenes; its scene set has zero overlap with the 561 training scenes.

\paragraph{Generation prompt.}
The teacher is instructed as follows: \emph{You will be shown four screenshots of the same target object from a ScanNet scene. In every image, the target is marked by a bounding box. Describe only its visible local appearance and small objects directly attached to, placed on, or supported by it. Report a one--two sentence target summary, visible details, attached or supported objects, parts, materials, state, local relations, and uncertain evidence. Exclude room-level and long-range spatial relations; do not infer invisible properties or guess. Use an empty list when a field is not visible and return strict JSON only.} The required keys are \texttt{target\_summary}, \texttt{visible\_details}, \texttt{attached\_or\_supported\_objects}, \texttt{parts}, \texttt{materials}, \texttt{state}, \texttt{local\_relations}, and \texttt{uncertain}.

\paragraph{Multitask composition and optimization.}
One epoch contains all 79,445 SQA3D, 26,515 ScanQA, 36,665 ScanRefer, and 43,838 Multi3DRefer examples, together with 23,560 detailed-caption examples; hence task sampling is proportional to these counts, with detailed captions comprising 11.2\% of the 210,023-example mixture. We use AdamW with a global batch size of 16, peak learning rate $10^{-5}$, vision-tower learning rate $2\times10^{-6}$, zero weight decay, 3\% warmup, and cosine decay. Training uses bfloat16, gradient checkpointing, and ZeRO-3 for 13,125 optimizer steps on eight NVIDIA H20 GPUs.

\paragraph{Updated modules and student input.}
We perform full-parameter tuning of the main vision--language stack rather than LoRA: the vision tower, multimodal projector, language model, spatial positional embeddings, and grounding head are updated, while the vision resampler remains frozen. The student uniformly samples 32 native scene views and applies the base models' SigLIP patch-14 preprocessing with 384-pixel base tiles and \texttt{anyres\_max\_9}; the four focused teacher views and wireframes are never exposed to the student.

\paragraph{Frame-quality score.}
Candidates are evaluated every eight frames and must have projected target-area ratio $r\geq0.002$. Let $c$ be the fraction of front-facing box corners inside the image, $i$ the fraction of the unclipped projected rectangle inside the image, $m$ its normalized minimum edge margin, $d$ the projected-center proximity to the image center, and $u$ indicate whether the projected target-region center lies inside the image. We define
\begin{align}
a(r)&=\begin{cases}
\min(r/0.12,1),&r\leq0.12,\\
1,&0.12<r\leq0.35,\\
\max(0,1-(r-0.35)/0.35),&r>0.35,
\end{cases}\\
f&=0.50i+0.35a(r)+0.15m,\qquad
g=0.20c+0.25f+0.30d+0.25u.
\end{align}
We multiply $g$ by 0.35 when $u=0$ and by 0.50 when the projected 3D box center is outside the image. For the best $40K=160$ geometric candidates, depth consistency uses tolerance $\tau=\max(0.3\,\mathrm{m},0.15z_{\min})$. Let $o$ be the fraction of ROI depth pixels closer than $z_{\min}-\tau$, $h$ the fraction within $[z_{\min}-\tau,z_{\max}+\tau]$, and $v=h(1-o)$. A candidate is depth-visible when $o<0.9$ and $h>0.15$, and its final quality is
\begin{equation}
s_t=(0.85g+0.15v)\,p(o),\qquad
p(o)=\begin{cases}1,&o\leq0.5,\\ \max(0,1-o^2),&o>0.5.\end{cases}
\end{equation}
If no candidate passes the depth test, geometric candidates are retained with $v=p=0.5$. Bbox-voxel then uses $K=4$, a 12-frame minimum gap, 12 voxel cells along the shortest box axis, depth subsampling by four, and a 2\% box margin.

\subsection{View-coverage comparison with ExCap3D}
\label{supp:view-coverage}

We compare the joint target-region coverage of LensUnd and the visibility-based view selection used by ExCap3D on all 2,068 LensBench targets. Both methods use the same budget of four views and the same voxelization. For each frame, depth pixels are unprojected into 3D, and only occupied voxels inside the target box are retained. Joint coverage is the number of unique voxels in the union of the four selected views. We also report inter-view redundancy,
\begin{equation}
1-\frac{\left|\bigcup_{t\in\mathcal{S}}\mathcal{V}_t\right|}
{\sum_{t\in\mathcal{S}}|\mathcal{V}_t|},
\end{equation}
where a lower value indicates less repeated evidence across views.

\begin{table}[ht]
\centering
\caption{Target-region view coverage under the same four-view budget. Union-normalized coverage uses, for each target, the union of voxels observed by either method as the reference set; it is not absolute full-surface coverage.}
\label{tab:view-coverage}
\begin{tabular}{lrr}
\toprule
Metric & LensUnd & ExCap3D \\
\midrule
Unique covered voxels & \textbf{2,057.7} & 1,564.6 \\
Union-normalized coverage (\%) & \textbf{90.3} & 63.6 \\
Inter-view redundancy (\%) $\downarrow$ & \textbf{38.0} & 47.8 \\
\bottomrule
\end{tabular}
\end{table}

LensUnd covers 31.5\% more unique voxels on average and obtains higher coverage on 97.1\% of targets, while reducing redundancy by 9.8 percentage points. ExCap3D ranks frames independently by visible-vertex fraction, so high-ranking frames may be temporally close and observe nearly identical target surfaces. In contrast, LensUnd greedily selects views according to their marginal voxel contribution, yielding more complementary evidence. Higher joint coverage exposes the target from a wider range of viewpoints, making otherwise occluded sides, parts, materials, and state cues more likely to be observed. It also uses the limited view budget more efficiently by reducing repeated evidence, thereby providing the VLM with a more complete basis for fine-grained target understanding.

\paragraph{Relation to ExCap3D.}
ExCap3D focuses on object- and part-level captioning with dedicated captioners, while Lens3D uses a shared target-focused evidence interface across fine-grained object captioning, fine-grained object QA, and small-object grounding. LensDistill further uses the resulting detailed captions to train pretrained generalist 3D LLMs together with their native grounding and scene-level QA data. The two approaches thus differ in task scope, model target, and view-selection strategy.

\subsection{Visual-dependence analysis}
\label{supp:visual-analysis}

\paragraph{Target-region masking.}
The student input contains no projected wireframe, only 32 RGB views and the \coord\ token encoding the target center. We project the ground-truth oriented 3D box into every RGB view and replace pixels inside its 2D enclosing rectangle with gray. This preserves scene context, the view sequence, and target-coordinate encoding while removing most target appearance. As shown in Table~\ref{tab:masking}, masking substantially reduces lexical, overall, and semantic description scores, confirming that generation depends materially on target pixels.

\begin{table}[ht]
\centering
\caption{Target-region masking for Lens3D (3DRS).}
\label{tab:masking}
\begin{tabular}{lrrr}
\toprule
Input & Summary F1 & Overall F1 & Summary Sem. \\
\midrule
Unmasked & 0.489 & 0.470 & 0.753 \\
Target masked & 0.369 & 0.337 & 0.560 \\
\bottomrule
\end{tabular}
\end{table}

\paragraph{Attention concentration and target alignment.}
We analyze attention from language-model query tokens to visual tokens during target-summary generation. Attention concentration is the mass assigned to the top 5\% of image patches. Target alignment is the ratio between attention within the projected target and its expectation under a uniform distribution. In the final four layers, the top 5\% of patches receive 35.0\% of answer-token attention on average, and every sample exceeds five times the uniform expectation.

Target alignment rises systematically with depth. Answer-token alignment increases from 1.29 times uniform expectation in layers L0--3 to approximately four to five times in L12--15 and 8.96 times in L24--27. The \coord\ token retains approximately 4.47 times target alignment in deep layers despite an attention-sink effect at non-target frame positions. The masking and attention results jointly show that LensDistill uses \coord\ to retrieve target appearance from the native multiview sequence.

\subsection{Attribute-level LensBench analysis}
\label{supp:attribute-analysis}
\enlargethispage{3\baselineskip}

\begin{table*}[t]
\centering
\caption{Full LensBench evaluation. ``Mean'' averages Parts, Materials, and State F1. Temporal and quality-pool denote uniform sampling from the full sequence and from a quality-filtered pool, respectively. Best and second-best results within each block are bolded and underlined, respectively.}
\label{tab:lensbench-full}
\scriptsize
\setlength{\tabcolsep}{2.5pt}
\resizebox{\textwidth}{!}{%
\begin{tabular}{lllp{2.35cm}ccrrrrrr}
\toprule
Type & Model & Box source & View selection & Views & Wire. &
Sum.\ F1 & Sum.\ Sem. & Parts & Materials & State & Mean \\
\midrule
LensUnd & Qwen3-VL-2B~\citep{bai2025qwen3} & Ground truth & Temporal uniform & 4  & No  & 0.263 & 0.619 & 0.192 & 0.607 & 0.272 & 0.357 \\
LensUnd & Qwen3-VL-2B & Ground truth & Temporal uniform & 4  & Yes & 0.370 & 0.694 & 0.218 & 0.591 & 0.275 & 0.362 \\
LensUnd & Qwen3-VL-2B & Ground truth & Temporal uniform & 32 & Yes & 0.380 & 0.686 & 0.201 & 0.589 & 0.278 & 0.356 \\
LensUnd & Qwen3-VL-2B & Ground truth & Quality-pool uniform & 4 & Yes & 0.426 & 0.732 & 0.245 & 0.628 & 0.300 & 0.391 \\
LensUnd & Qwen3-VL-2B & Ground truth & Video-3D LLM greedy & 4 & Yes & 0.353 & 0.690 & 0.215 & 0.594 & 0.277 & 0.362 \\
LensUnd & Qwen3-VL-2B & Ground truth & ExCap3D visibility & 4 & Yes & 0.436 & 0.742 & 0.250 & \underline{0.634} & 0.299 & 0.394 \\
LensUnd & Qwen3-VL-2B & Ground truth & bbox-voxel & 4 & Yes & \underline{0.445} & \underline{0.747} & \underline{0.252} & 0.632 & \underline{0.308} & \underline{0.397} \\
LensUnd & Qwen3-VL-2B & 3DRS & bbox-voxel & 4 & Yes & 0.435 & 0.739 & 0.244 & 0.625 & 0.305 & 0.391 \\
LensUnd & Qwen3-VL-2B & Video-3D LLM & bbox-voxel & 4 & Yes & 0.434 & 0.736 & 0.249 & 0.630 & 0.302 & 0.393 \\
LensUnd & Qwen3-VL-8B & Ground truth & bbox-voxel & 4 & Yes &
\textbf{0.485} & \textbf{0.783} & \textbf{0.474} & \textbf{0.726} & \textbf{0.384} & \textbf{0.528} \\
\midrule
3D LLM & 3DRS & Ground truth & Temporal uniform & 32 & No & 0.230 & 0.549 & -- & -- & -- & -- \\
3D LLM & \textbf{Lens3D (3DRS)} & Ground truth & Temporal uniform & 32 & No &
\textbf{0.489} & \textbf{0.752} & \textbf{0.477} & \textbf{0.709} & \textbf{0.429} & \textbf{0.538} \\
3D LLM & Video-3D LLM & Ground truth & Temporal uniform & 32 & No & 0.228 & 0.542 & -- & -- & -- & -- \\
3D LLM & \textbf{Lens3D (Video-3D LLM)} & Ground truth & Temporal uniform & 32 & No &
\underline{0.483} & \underline{0.747} & \underline{0.473} & \underline{0.698} & \underline{0.428} & \underline{0.533} \\
\bottomrule
\end{tabular}}
\end{table*}

In addition to summary quality, Table~\ref{tab:lensbench-full} reports lexical set agreement for parts, materials, and state. We use multi-reference mean set-token F1: each predicted field is compared independently with each of the three references, after which results are averaged over references and samples. The original 3DRS, Video-3D LLM, Chat-3D v2, and ChatScene outputs are free-form and do not provide corresponding attribute fields. We therefore omit their attribute scores rather than compare full-response tokens against structured fields under a different extraction procedure.

After LensDistill, the three-field means reach 0.538 and 0.533 for 3DRS and Video-3D LLM, respectively. Under identical LensUnd evidence, Qwen3-VL-8B improves over the 2B model by 22.2, 9.4, and 7.6 percentage points on parts, materials, and state; its Materials F1 of 0.726 is the highest in the table. Scores remain similar across box sources. Increasing temporal sampling from four to 32 views reduces the three-field mean from 0.362 to 0.356, consistent with the summary results.

\end{document}